\documentclass{egpubl}
 
\JournalSubmission    
\usepackage[T1]{fontenc}
\usepackage{dfadobe}  

\usepackage{cite}  
\BibtexOrBiblatex
\electronicVersion
\PrintedOrElectronic
\ifpdf \usepackage[pdftex]{graphicx} \pdfcompresslevel=9
\else \usepackage[dvips]{graphicx} \fi

\usepackage{egweblnk} 

\usepackage{tikz}
\usetikzlibrary{arrows,shapes,positioning,shadows,trees}
\usepackage{forest}
\usepackage{tikz-qtree}
\usetikzlibrary{arrows.meta}
\usepackage[export]{adjustbox}
\usepackage{float} 
\usepackage{booktabs}
\usepackage{multirow}
\usepackage{tabularx}
\usepackage{diagbox}

\usepackage{amsmath}

\newcommand{\etal}{{et al}. }

\usepackage{amsfonts}

\usepackage{mathtools} 
\usepackage{xspace} 
\newcommand{\FLIP}{\protect\reflectbox{F}LIP\xspace}

\usepackage{url}

\usepackage{colortbl}

\title[PQDAST: Arbitrary Style Transfer for Games]%
      {PQDAST: Depth-Aware Arbitrary Style Transfer for Games \\ via Perceptual Quality-Guided Distillation}

\author[E. Ioannou \& S. Maddock]
{\parbox{\textwidth}{\centering E. Ioannou\orcid{0000-0003-3892-2492}
        and S. Maddock\orcid{0000-0003-3179-0263} 
        }
        \\
{\parbox{\textwidth}{\centering The University of Sheffield, UK}
}
}

\begin{document}


\teaser{
 \includegraphics[width=\textwidth]{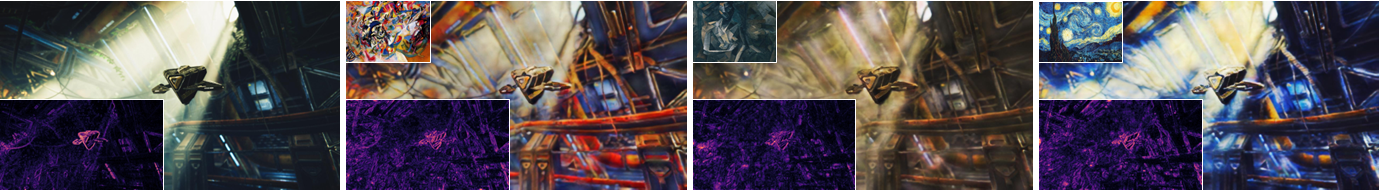}
 \centering
  \caption{In-game artistic stylisations generated using PQDAST. The original input is shown on the left. To visualise the efficiency of our approach in achieving temporal coherence, we compute the differences between the shown frame and the previous frame using the \FLIP evaluator (bottom left for each frame). The temporal error difference values as calculated using \FLIP (lower is better) are 0.0742 (original), 0.1021, 0.1281, 0.1219 from left to right.}
    \label{fig:teaser}
}

\maketitle

\begin{abstract}
    Artistic style transfer is concerned with the generation of imagery that combines the content of an image with the style of an artwork. In the realm of computer games, most work has focused on post-processing video frames. Some recent work has integrated style transfer into the game pipeline, but it is limited to single styles. Integrating an arbitrary style transfer method into the game pipeline is challenging due to the memory and speed requirements of games. We present PQDAST, the first solution to address this. We use a perceptual quality-guided knowledge distillation framework and train a compressed model using the \FLIP evaluator, which substantially reduces both memory usage and processing time with limited impact on stylisation quality. For better preservation of depth and fine details, we utilise a synthetic dataset with depth and temporal considerations during training. 
    The developed model is injected into the rendering pipeline to further enforce temporal stability and avoid diminishing post-process effects. 
    Quantitative and qualitative experiments demonstrate that our approach achieves superior performance in temporal consistency, with comparable style transfer quality, to state-of-the-art image, video and in-game methods.



\end{abstract}

\section{Introduction}
\label{sec:Introduction}

Arbitrary Neural Style Transfer (NST) uses the style of any artwork to alter content data such as images \cite{huang2017arbitrary,park2019arbitrary,liu2021adaattn,ma2023rast}, videos \cite{li2019learning,deng2021arbitrary,lu2022universal} and radiance fields \cite{zhang2022arf,liu2023stylerf,pang2023locally,li2024s,fischer2024nerf}. In the realm of computer games, image and video NST methods can be utilised as post-processing effects at the end of a game's rendering pipeline. Nevertheless, this treats artistic stylisation as a final filter, ignoring the 3D nature of a computer game's scene, which can result in temporal instabilities and undesired flickering effects. Recently, work has focused on artistic style transfer specifically tailored for games \cite{ioannou2023games,ioannou2024towards}, but it is constrained to a single style. 

The ability to arbitrarily stylise 3D scenes could be a potent tool for game development. However, a challenge is to maintain high stylisation quality while addressing inherent speed and memory issues. Using intermediate (G-buffer) information that becomes available during the rendering process shows promise for addressing this.
Recent methods have demonstrated remarkable improvements in the quality of generated stylised game scenes using G-buffer data \cite{richter2022enhancing,Mittermueller2022estgan}, while other approaches have integrated NST as part of the 3D computer graphics pipeline to alleviate the issue of temporal incoherence across subsequent frames \cite{ioannou2023games,ioannou2024towards}. These methods avoid applying a trained image or video style transfer approach at the end of the rendering process, however, they utilise a conventional convolutional-based transformation network that is only capable of reproducing one artistic style.

\begin{figure}[ht]
  \centering
    \includegraphics[width=\linewidth]{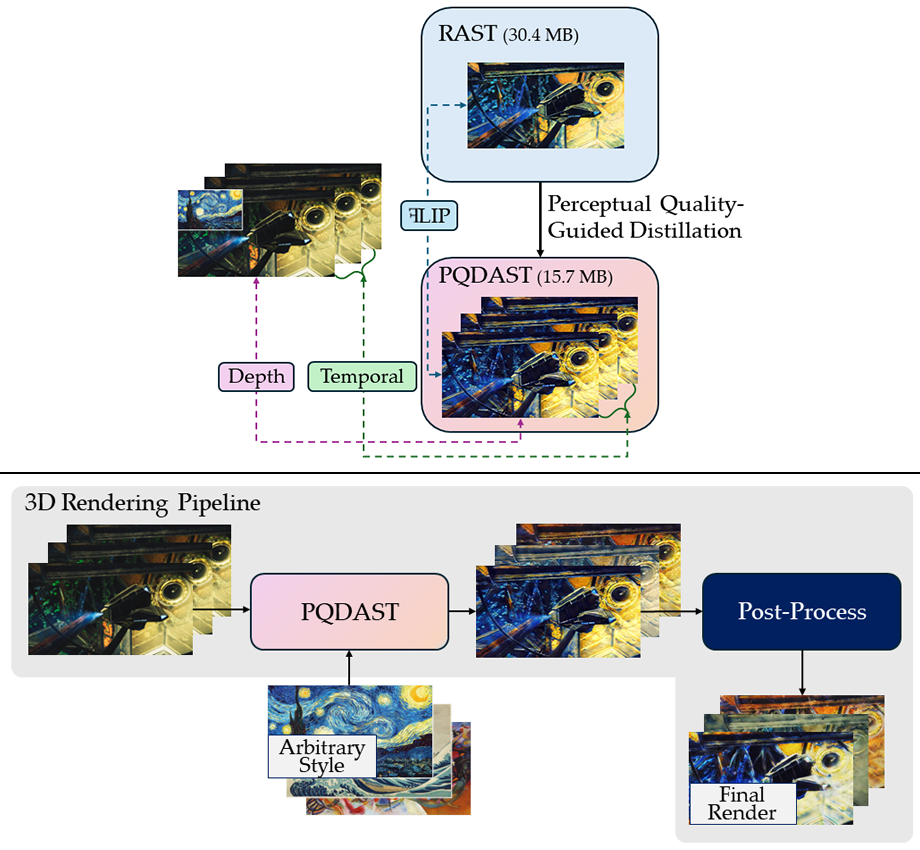}
   \caption{Our proposed perceptual quality-guided knowledge distillation framework utilises the \FLIP evaluator. Depth and temporal loss functions are also defined. The trained model is injected into the 3D rendering pipeline.}
   \label{fig:system_overview}
\end{figure}


In this paper, we introduce \textit{PQDAST}, which, to our knowledge, is the first arbitrary style transfer approach that is injected in the game's rendering pipeline \cite{ioannou2023games,ioannou2024towards}.
In most cases, algorithms that are capable of reproducing an arbitrary style per trained network \cite{park2019arbitrary,liu2021adaattn,ma2023rast} first extract image features of content and style and then do a forward pass through a trained transformer network before a decoder generates the stylised result. We propose a solution that is based on the approach of Ma~\etal \shortcite{ma2023rast}, which involves training a compressed transformer and decoder network using knowledge distillation. We devise a new loss function that is inspired by work on image quality assessment \cite{Andersson2020} to force the compressed models to retain the quality of the original model. Our novel perceptual quality-guided distillation loss illustrates how image quality assessment research can contribute to model compression for effectively minimising the speed and memory of algorithms dedicated to image generation. Additionally, we utilise an advanced depth estimation network \cite{yang2024depth} for a depth reconstruction loss that was previously shown to improve the quality of style transfer results \cite{liu2017depth,ioannou2022depth}. Instead of training on an image dataset, we train on a synthetic video dataset and employ a temporal loss function for improved temporal stability. Figure~\ref{fig:teaser} provides an overview of our proposed method. The contributions of our work can be summarised as follows:

\begin{itemize}
     \item We propose a solution for arbitrary style transfer in computer games, enabling users to apply any painting to artistically alter the visuals of the game. 

    \item We present a technique that compresses the model of a previous approach \cite{ma2023rast} to approximately half its size. Our algorithm utilises the \FLIP evaluator in a novel perceptual quality-guided knowledge distillation technique that achieves comparable stylisation quality and improved temporal stability compared to state-of-the-art methods.

    \item Our developed network is integrated into the computer game's rendering pipeline (similar to \cite{ioannou2023games,ioannou2024towards}) resulting in enhanced temporal consistency.

    \item Extensive qualitative and quantitative experiments demonstrate that our proposed algorithm achieves high-quality arbitrary style transfer for computer games.
    
\end{itemize}

\section{Related Work}

\subsection{Image \& Video Style Transfer}

NST research has progressed from online image-optimization techniques \cite{gatys2016image}, to offline model-optimization methods capable of reproducing one style per trained network \cite{johnson2016perceptual,li2016precomputed,ulyanov2016texture,ulyanov2017improved}, and arbitrary-style-per-model approaches \cite{huang2017arbitrary} that can reproduce any given referenced style image on an input photograph \cite{ghiasi2017exploring,gu2018arbitrary,huo2021manifold,shen2018neural,svoboda2020two,xu2023learning}. Early work on arbitrary style transfer, \textit{AdaIN} \cite{huang2017arbitrary}, used an adaptive instance normalization layer that aligns the mean and variance of the content features with the respective mean and variance of style features. Other work suggested patch-based techniques \cite{chen2016fast,sheng2018avatar}, while neural flows \cite{an2021artflow} and vector quantization \cite{huang2023quantart} have also been exploited for arbitrary stylisation. Recently, the success of attention mechanisms \cite{vaswani2017attention,dosovitskiy2020image} in computer vision has resulted in multiple attention-based methods \cite{park2019arbitrary,liu2021adaattn,deng2022stytr2,luo2022consistent,ma2023rast,hong2023aespa,zhu2023all}, as well as diffusion model-based methods \cite{chung2024style,hertz2024style} for artistic style transfer. Among these attention-based approaches, \textit{RAST} \cite{ma2023rast}, a system inspired by image restoration shows enhanced structure preservation, a desirable quality in a game setting. Our approach utilises \textit{RAST} (which uses \textit{SANet} \cite{park2019arbitrary} as a backbone) in a distillation framework that is also based on style-attentional networks (\textit{SANet}).


Temporal incoherence is the main challenge that arises when stylising videos compared to images. Methods have resorted to optic flow data to improve temporal stability \cite{ruder2016artistic,gao2018reconet}. Multiple-style-per-network models \cite{gao2020fast} and arbitrary-style-per-network models \cite{deng2021arbitrary,lu2022universal} have been proposed, while depth-aware and structure-preserving video style transfer \cite{cheng2019structure,liu2021structure,ioannou2023depth} attempts to retain depth and global structure of the stylised video frames. Image style transfer approaches have been extended to work for videos with additional temporal loss training \cite{li2019learning,liu2021adaattn}, and unified frameworks for joint image and video style transfer techniques have been developed \cite{gu2023two,zhang2023unified}. Diffusion-based methods for stylised video generation have also emerged \cite{ku2024anyv2v}.

\subsection{Style Transfer for 3D Computer Games}

Whilst image and video NST methods can be applied at the end of the rendering pipeline to achieve real-time computer game stylisation, this is essentially a post-processing effect that interprets the rendered frames as single images and does not prevent undesired artifacts and flickering issues. Multi-style artistic style transfer for games has been shown in work by Unity \cite{deliot_guinier_vanhoey_2020} -- this utilises the method of Ghiasi~\etal \shortcite{ghiasi2017exploring} to stylise each intercepted final rendered image. Any G-buffer or 3D data is ignored while the produced stylisations are inconsistent and the post-process effects are diminished, as the stylisation network is used as a final `filter'. Other approaches have demonstrated improved stylisation quality when G-buffer data is taken into account during training \cite{richter2022enhancing,Mittermueller2022estgan}. Style transfer specifically tailored for computer games has only been recently proposed \cite{ioannou2023games,ioannou2024towards}. Here, NST is injected into the rendering pipeline before the post-process stage but is only capable of reproducing one style image per trained network. Yet, arbitrary style transfer could offer a significant advantage to developers and artists, as well as enable users to upload any artwork of their choice to stylise the game scenes.

\subsection{Knowledge Distillation}
Pioneered by Hinton~\etal \shortcite{hinton2015distilling}, knowledge distillation has been a widely adopted technique for training compressed models. This aims to create smaller and faster models that retain quality and performance. Recently, methods have leveraged this technique for the task of style transfer, demonstrating improved performance \cite{wang2020collaborative,chen2020optical,chen2023kbstyle}. Wang~\etal \shortcite{chen2023collaborative} show that training a smaller encoder to replace the large \textit{VGG-19} \cite{simonyan2015deep} that is typically utilised in encoder-decoder-based neural style transfer results in ultra-resolution outputs that were hard to achieve before due to memory constraints. High-quality arbitrary style transfer for images is also achieved by designing a network composed of a content encoder, a style encoder and a decoder based on CNNs, and employing symmetric knowledge distillation \cite{chen2023kbstyle}. The method by Chen~\etal \shortcite{chen2020optical} -- also based on a simple CNN architecture -- achieves fast video style transfer without relying on optic flow information during inference, but is only capable of reproducing one style per trained network.

\section{Our Approach}

\begin{figure*}[htb]
  \centering
   \includegraphics[width=\linewidth]{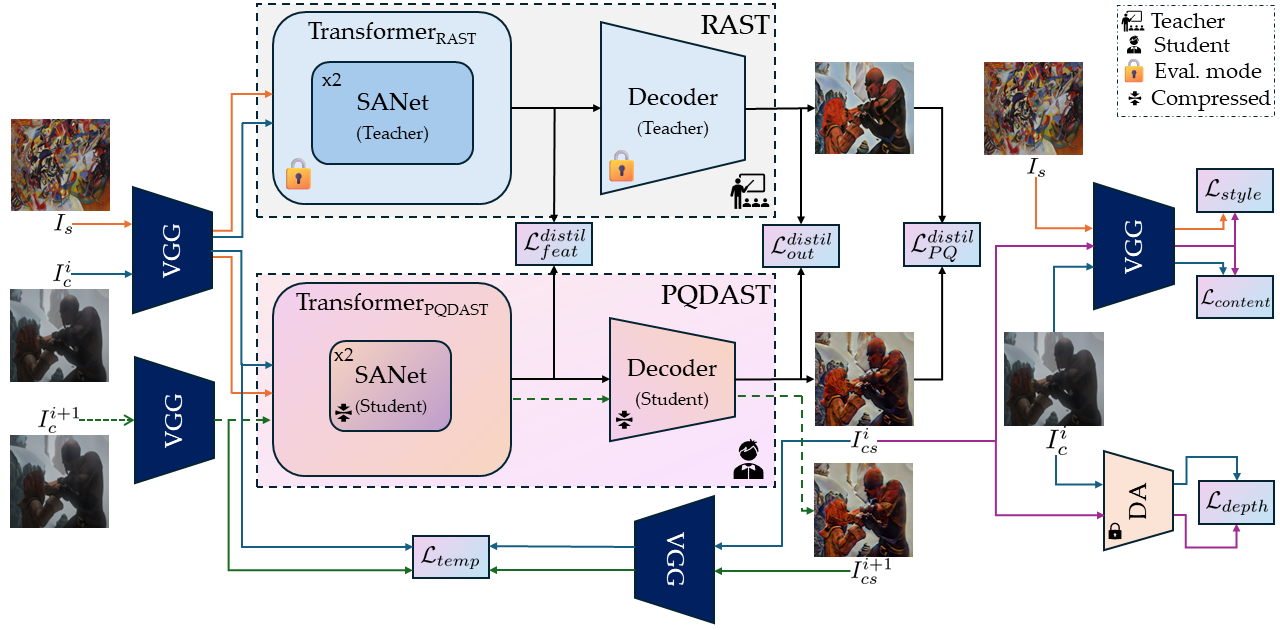}
   \caption{Overview of PQDAST Architecture. PQDAST trains a compressed version of RAST's decoder and SANet modules using perceptual quality-guided distillation losses. It also uses a depth reconstruction loss and a temporal loss in addition to the content and style losses.}
   \label{fig:system}
\end{figure*}

Figure~\ref{fig:system} provides an overview of the proposed system architecture. We adopt a widely used encoder-decoder design; we train a transformer that encompasses compressed versions of the SANet module, and a small decoder to produce comparable results to \textit{RAST} that uses \textit{SANet} as a backbone. In addition to distillation losses, we employ a temporal loss \cite{liu2021adaattn} and a depth reconstruction loss \cite{liu2017depth,ioannou2022depth,ioannou2023games} for stable and high-quality results.

\subsection{Preliminaries: \textit{SANet}, \textit{RAST}}

The recent technique by Ma~\etal \shortcite{ma2023rast,ma2024rast}, based on \textit{SANet} \cite{park2019arbitrary}, has shown remarkable performance in terms of alleviating the Content Leak phenomenon. \textit{RAST} \cite{ma2023rast}, due to its image restoration capabilities, achieves a high perceptual similarity score which means that fine details are efficiently preserved. This is desirable in the context of computer games. 
In addition, \textit{RAST} utilises two external discriminators to handle realistic-to-artistic and artistic-to-realistic processes. This ability of the model to stylise images in both directions, combined with its capability to handle photorealistic as well as artistic style transfer, makes it suitable for games that may feature a non-photorealistic style or strive for photorealism.
We therefore design a technique that distils the knowledge of \textit{RAST} to a compressed model. 

\textit{RAST} uses \textit{SANet} as a backbone. Assuming content image $I_c$ and style image $I_s$, and given encoded content and style feature maps $F_c$ and $F_s$, from a pre-trained \textit{VGG} \cite{simonyan2015deep}, the SANet module transforms them into two feature spaces $f$ and $g$ and calculates the attention between $\overline{F^{i}_c}$ and $\overline{F^i_s}$, where $\overline{F}$ denotes mean-variance channel-wise normalised version of F:
\begin{equation}
    F^i_{cs} = \frac{1}{C(F)} \sum_{\forall j} exp(f(\overline{F^{i}_c})^T g(\overline{F^j_s})) h(F^j_s),
\end{equation}
where $f(\overline{F_{c}}) = W_f\overline{F_{c}}$, $g(\overline{F_{s}}) = W_g\overline{F_{s}}$ and $h(F_{s}) = W_hF_{s}$ are learned weight matrices implemented as $1 \times 1$ convolutions.

This output feature map is then used to obtain $F_{csc}$:
\begin{equation}
    F_{csc} = F_c + W_{cs}F_{cs}
\end{equation}

Two SANet modules are used for features extracted from layers $relu4\_1$, and $relu5\_1$ of \textit{VGG}, respectively. The outputs of the two SANets are then combined:
\begin{equation}
    F^m_{csc} = conv_{3\times3} (F^{relu4\_1}_{csc} + upsampling(F^{relu5\_1}_{csc})),
\end{equation}
before the decoder synthesises the final output:
\begin{equation}
    I_{cs} = D(F^m_{csc}).
\end{equation}

\begin{table}[htb]
\begin{center}
    \caption{The network architecture of the original Style-Attentional Network (SANet) compared to our proposed compressed SANet used in PQDAST.}
    \label{tab:network_architecture}
        \begin{tabular}{l c  c}
           \toprule
             & \textbf{SANet} & \textbf{PQDAST}  \\
            \midrule
            Layer & \multicolumn{2}{c}{Features In $\rightarrow$ Features Out} \\ \hline

            Conv (f) & 512 $\rightarrow$ 512 & 512 $\rightarrow$ 256 \\ \hline 
            Conv (g) & 512 $\rightarrow$ 512 & 512 $\rightarrow$ 256 \\ \hline 
            Conv (h) & 512 $\rightarrow$ 512 & 512 $\rightarrow$ 256 \\ \hline 
            Conv (out) & 512 $\rightarrow$ 512 & 256 $\rightarrow$ 512 \\ \bottomrule 
            \end{tabular}
\end{center}
\end{table}

\begin{figure}[ht]
  \centering
   \includegraphics[width=\linewidth]{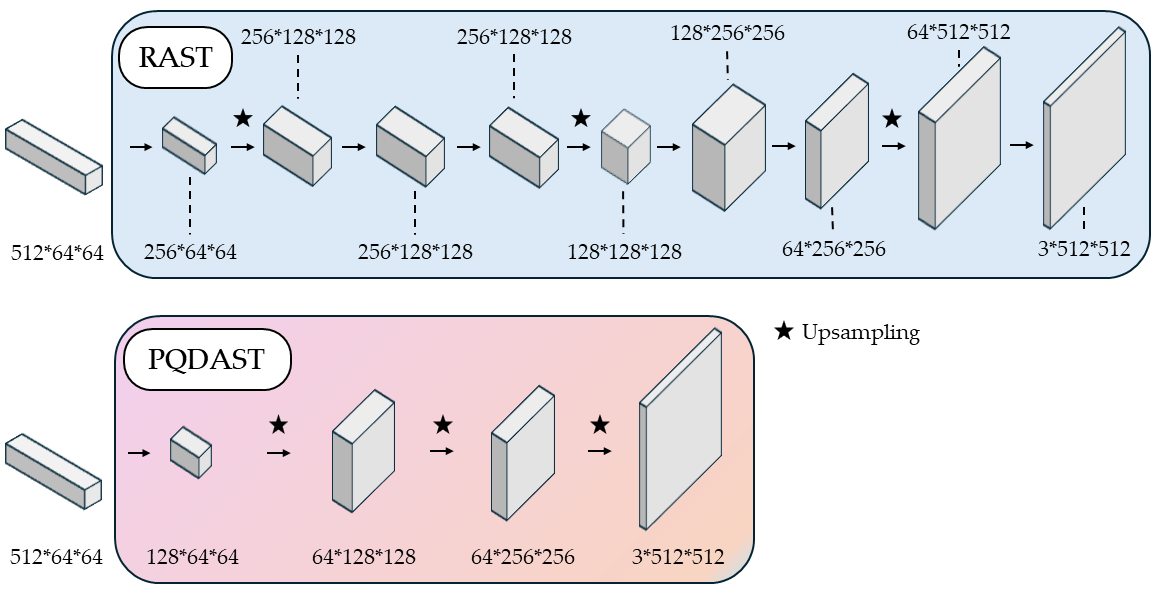}
   \caption{Decoder Architecture: \textit{RAST} vs \textit{PQDAST}.} 
   \label{fig:decoder_architecture}
\end{figure}

Our neural network architecture resembles the architecture of \textit{SANet} but it has significantly reduced complexity. As the SANet module is used twice, we define a student transformer network that utilises a student SANet module with reduced feature maps of each convolution layer, as shown in Table~\ref{tab:network_architecture}. This reduces the floating point operations performed (FLOPs) from 15.05G to 12.35G and the number of parameters from 4.46M to 3.41M for the transformer block that combines the outputs of the two SANet modules. In addition, we compress the decoder network from 9 convolutional layers (63.36G FLOPs, 3.51M parameters) to 4 convolutional layers (6.51G FLOPs, 702.40K parameters), as illustrated in Figure~\ref{fig:decoder_architecture}.

\subsection{Perceptual Quality-Guided Knowledge Distillation}
\label{sec:distillation}

Our proposed framework is trained using a combination of three distillation losses. As our transformer is less complex than the transformer used in \textit{RAST}, combining the outputs of two SANet modules, we initially define a loss that minimises the error between the outputs of the \textit{RAST}'s transformer ($F^m_{csc_{teacher}}$) and \textit{PQDAST}'s transformer ($F^m_{csc_{student}}$):
\begin{equation}
    \mathcal{L}_{feat}^{distil} = \| F^m_{csc_{student}} -  F^m_{csc_{teacher}} \|^{2}_{2}
\end{equation}

We repeat the same for the outputs of the decoders:
\begin{equation}
    \mathcal{L}_{out}^{distil} = \| I_{cs_{student}} -  I_{cs_{teacher}} \|^{2}_{2}
\end{equation}

The plethora of image and video style transfer methods do not optimise for applicability in computer games. In addition to temporal considerations that are useful for games, we also utilise a synthetic video dataset to better capture the synthetic nature of the visual media our algorithm intends to stylise. Nevertheless, it is necessary for our approach to remain perceptually consistent to \textit{RAST}, which achieves good results in terms of sustaining fine details. Training on a different dataset (and with a less complex model) would result in dissimilarities between the outputs. To better match the output of \textit{RAST}, we define a distillation loss based on perceptual quality. For this, we treat the output of our model as a `rendered image' that is an attempt to reproduce the output of \textit{RAST}, which can be considered as the `ground truth'. 

For this task, we use \FLIP, as SSIM has been shown to lack the necessary qualities for use with colour images \cite{nilsson2020understanding}. 
\FLIP's architecture is based on a colour pipeline and a feature pipeline, resulting in an image quality metric that performs competently against state-of-the-art methods and coincides with human judgement. %
The specific focus of \FLIP on rendering quality makes it particularly well-suited for our method.  While many image quality metrics are designed for general image comparison, \FLIP is tailored to assess the types of artefacts and differences commonly encountered in rendered images. This specialization is essential for our distillation loss, as it allows us to penalize precisely those visual discrepancies that are most likely to be noticed by viewers. This targeted approach ensures that our distilled model learns to prioritize the aspects of image quality most relevant to rendering, leading to more visually compelling results.
We, thus, utilise \FLIP to define an additional perceptual quality-guided distillation loss:

\begin{equation}
    \mathcal{L}_{PQ}^{distil} = \FLIP(I_{cs_{student}},  I_{cs_{teacher}}),
\end{equation}

with the resulting total distillation loss defined as:

\begin{equation}
    \mathcal{L}_{total}^{distil} = \mathcal{L}_{feat}^{distil} + \mathcal{L}_{out}^{distil} + \mathcal{L}_{PQ}^{distil}.
\end{equation}

\subsection{Depth and Temporal Considerations}

Similarly to previous techniques optimised for computer games \cite{ioannou2023games,ioannou2024towards}, we employ a depth reconstruction loss to reinforce the retainment of depth in the synthesised results -- depth information has been consistently shown to enhance the quality of artistically stylised imagery \cite{liu2017depth,ioannou2022depth}. Unlike previous methods \cite{ioannou2023games,ioannou2024towards}, we adopt the recent method of Yang~\etal \shortcite{yang2024depth} (``\textit{Depth Anything}", here, denoted as \textit{DA}) which demonstrates improved performance compared to \textit{MiDaS} \cite{Ranftl2020}. The depth reconstruction loss is thus formulated as:

\begin{equation} \label{eq:depth-loss}
    \mathcal{L}^{DA}_{depth} (I_{cs}, I_c)  =  \| DA(I_{cs}) - DA(I_c) \|^{2}_{2}.
\end{equation}

Additionally, the proposed system, trained on synthetic video data, allows for temporal considerations. Our temporal loss ($\mathcal{L}_{temp}$) is adopted from Liu~\etal \shortcite{liu2021adaattn}. 



\subsection{Full System}
The overall loss function our system optimises is a weighted summation of the knowledge distillation loss $\mathcal{L}^{distil}_{total}$, depth loss $\mathcal{L}^{DA}_{depth}$, temporal loss $\mathcal{L}_{temp}$ and perceptual (content $\mathcal{L}_{content}$ and style $\mathcal{L}_{style}$) losses:
\begin{equation}
    \mathcal{L}_{total} = \lambda_c \mathcal{L}_{content} + \lambda_s \mathcal{L}_{style}  + \lambda_k \mathcal{L}^{distil}_{total} + \lambda_d \mathcal{L}^{DA}_{depth} + \lambda_t \mathcal{L}_{temp} 
\end{equation}

where content losses are adopted from \textit{SANet} \cite{park2019arbitrary}. Content loss is defined as:
\begin{equation} \label{eq:contentloss}
   \mathcal{L}_c = \| \overline{E(I_{cs})^{u}} - \overline{F_c^{u}} \|_2 + \| \overline{E(I_{cs})^{v}} - \overline{F_c^{v}} \|_2.
\end{equation}

with $u=relu4\_1$ and $v=relu5\_1$ layers of a pre-trained \textit{VGG-19} \cite{simonyan2015deep}, $\overline{F_c^{*}}$ denotes mean-variance channel-wise normalised content features, and $\overline{E(I_{cs})^{*}}$ denotes the corresponding mean–variance channel-wise normalised features of the stylised image. Style loss is defined as:

\begin{equation} \label{eq:style-loss}
    \mathcal{L}_{style} = \sum_{i=1}^{L} \left\| \mu(\phi_i(I_{cs})) - \mu(\phi_i(I_s)) \right\|_2 + \left\| \sigma(\phi_i(I_{cs})) - \sigma(\phi_i(I_s)) \right\|_2.
\end{equation}
where $L=\{relu1\_1, relu2\_1, relu3\_1, relu4\_1, relu5\_1\}$, and where $\phi_{i}$ denotes a feature map of the $i$-th layer of the \textit{VGG} encoder.

\subsection{PQDAST in the Game's Pipeline}

Inspired by previous work for in-game artistic stylisation \cite{ioannou2023games,ioannou2024towards}, we implement a \textit{Custom Pass} in the Unity HDRP \cite{unity_hdrp}. The trained network is injected before the Post-Process stage. The user can select any artwork to be used as the reference style image. This leads to generated results of improved temporal coherence for any selected style image, while the post-process effects (e.g., Depth-of-Field) are retained. It is important to note that our framework is trained with gamma-encoded images, whereas Unity HDRP uses a Linear colour space. Therefore, the reference style image and each colour buffer mipmap frame are converted to sRGB space before being processed.

\section{Experiments}

\subsection{Training Details}

Considering the synthetic nature of computer games' imagery, we use the \textit{MPI Sintel} \cite{Butler2012sintel} training dataset to train our network. As the trained \textit{PQDAST} is injected before the post-process stage, and intercepts frames that are not post-processed, we train using frames from both the Clean pass and the Final pass. \textit{Wikiart} \cite{phillips2011wiki} is used as the style images dataset. Adam optimizer \cite{kingma2014adam} is employed with a learning rate of $0.0001$ and a batch size of 6 content–style image pairs. During training, both images are rescaled to $256 \times 256$ pixels. The hyperparameters $\lambda_c$, $\lambda_s$, $\lambda_k$, $\lambda_d$, and $\lambda_t$ are set to $1.0$, $3.0$, $1.0$, $1.0$, and $10.0$ respectively. Training requires 160000 steps and lasts about 30 hours on a single NVIDIA Tesla V100 GPU.


\begin{figure*}[ht]
  \centering
   \includegraphics[width=\linewidth]{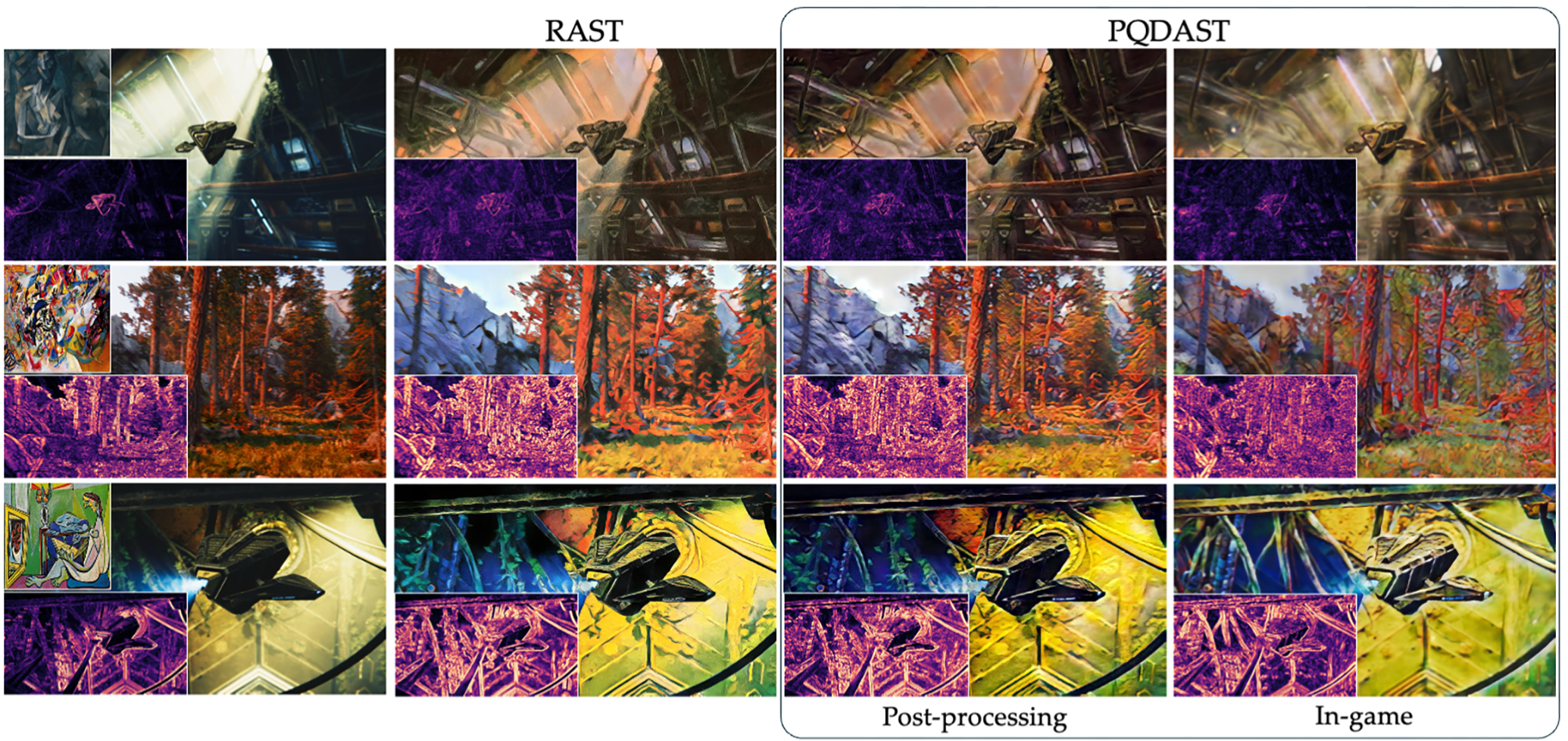}

   \hfill
    \centering

    \vspace{-0.2cm}
    
    \begin{tabular}{>{\centering\arraybackslash}m{3.7cm} >{\centering\arraybackslash} m{3.7cm}  >{\centering\arraybackslash} m{3.7cm} >{\centering\arraybackslash}m{3.7cm}  }
    \toprule
    \textbf{Input} & \textbf{RAST}  & \textbf{PQDAST (Post-processing)} & \textbf{PQDAST (in-game)} \\
    \midrule
      0.0742 & 0.1546 & 0.1354 & 0.1021  \\
      \hline
      0.3517 & 0.4969 & 0.4769 & 0.4110 \\
      \hline
      0.2445 & 0.4299 & 0.4248 & 0.4262 \\ 
      \bottomrule
    \end{tabular}

    \caption{Results comparing \textit{PQDAST} to \textit{RAST}. \textit{RAST} is used as a post-processing effect. The input frame is shown on the left. The difference between the shown and previous frames is visualised using the \FLIP evaluator. In-game \textit{PQDAST} generates temporally consistent results. The difference between the shown and previous frames is visualised using the \FLIP evaluator. The table below the images provides the numerical value of this difference (calculated using \FLIP), for each method and for each row of the figure.}
   
   \label{fig:results_ours_rast_2}
\end{figure*}

\subsection{PQDAST for Computer Games}

Our proposed framework trains compressed transformer and decoder models to generate results with comparable stylisation quality to \textit{RAST} \cite{ma2023rast}. Similarly to \cite{ioannou2023games,ioannou2024towards}, the trained model is injected into the game's pipeline. Example results are shown in Figure~\ref{fig:results_ours_rast_2}. At the bottom of each image, temporal error maps are provided (\FLIP is used to compute the difference between the current and previous frame). Our system used as a post-processing effect synthesises similar results to \textit{RAST}, with improved temporal consistency. When \textit{PQDAST} is used in-game, stylised frames are temporally more stable (the temporal error map is the closest in similarity with the original frame's temporal error map), while the stylisation quality is slightly altered, as the post-process effects in the game are enabled. In addition to the visual fidelity improvements introduced in \textit{PQDAST} and the temporal considerations we make during training, our system gains a boost in performance when injected into the game's pipeline. Intercepting each G-buffer colour frame and producing a stylised version that is then passed through the Post-process stage prevents undesired artefacts and flickering effects, as shown in \cite{ioannou2023games,ioannou2024towards}. Additional results of \textit{PQDAST} in-game are shown in Figure~\ref{fig:teaser}. 

\subsection{Comparisons with State-of-the-Art Methods}
We compare the performance of our approach against seven state-of-the-art methods. As temporal stability is crucial for the stylisation of computer games, we compare against approaches that consider temporal information (\textit{AdaAttN} \cite{liu2021adaattn}, \textit{CSBNet} \cite{lu2022universal}, \textit{MCCNet} \cite{deng2021arbitrary}, \textit{FVMST} \cite{gao2020fast}) or they are optimised for computer games (\textit{NSTFCG} \cite{ioannou2023games}, \textit{GBGST} \cite{ioannou2024towards}). We also compare against \textit{RAST} \cite{ma2023rast}, the method which our model distils knowledge from. 

\begin{figure*}[ht]
  \centering
   \includegraphics[width=\linewidth]{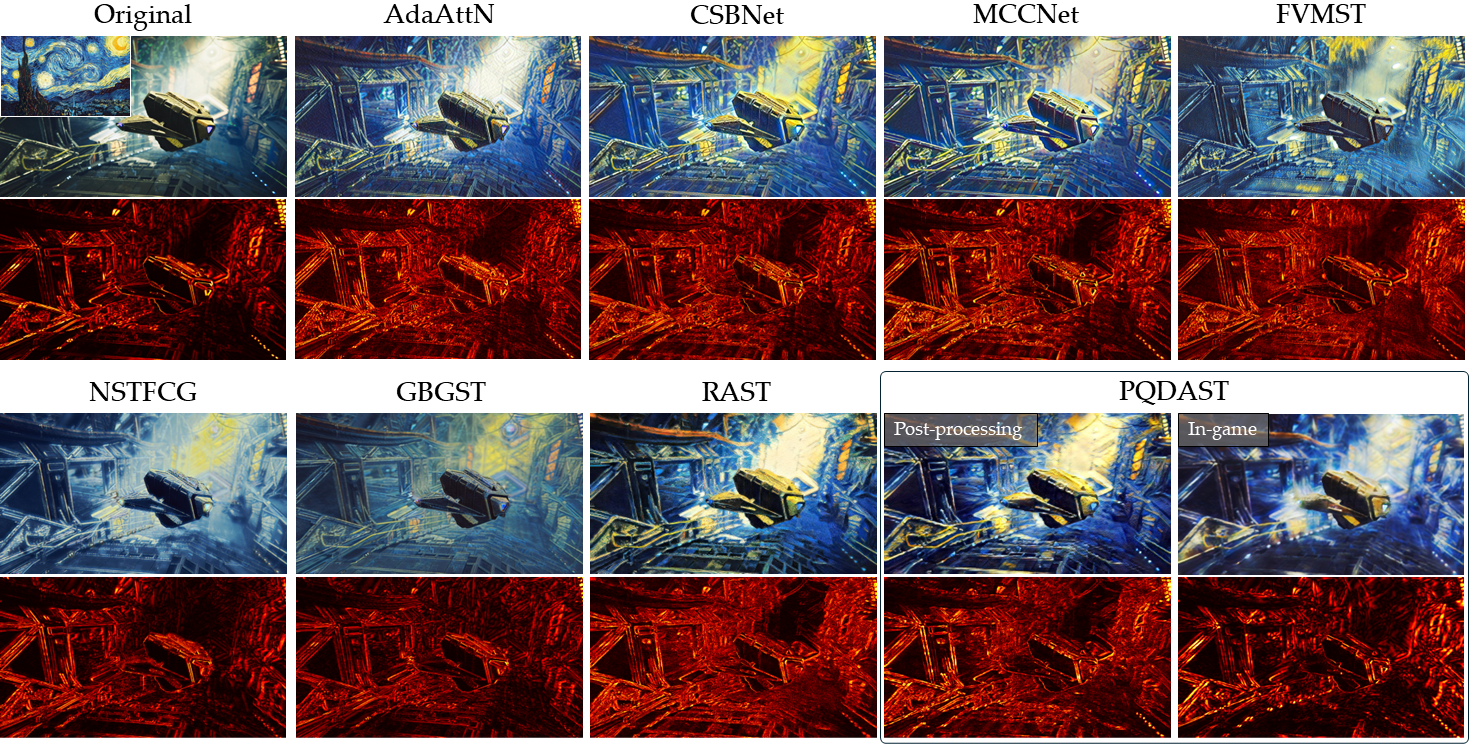}

      
     
    
   \caption{Qualitative results comparing \textit{PQDAST} to state-of-the-art methods. A heatmap of the temporal error between the current and previous frame is included in the bottom row. Our proposed approach produces high-quality stylisations. The temporal error heatmap of \textit{PQDAST} in-game is closest to the original frame's heatmap along with \textit{NSTFCG} and \textit{GBGST} that are used in-game. Additional results are provided in Figure~\ref{fig:results_comparison_2}. }
   \label{fig:qualitative}
\end{figure*}

\subsection{Qualitative Results}

\begin{figure*}[ht]
  \centering
   \includegraphics[width=\linewidth]{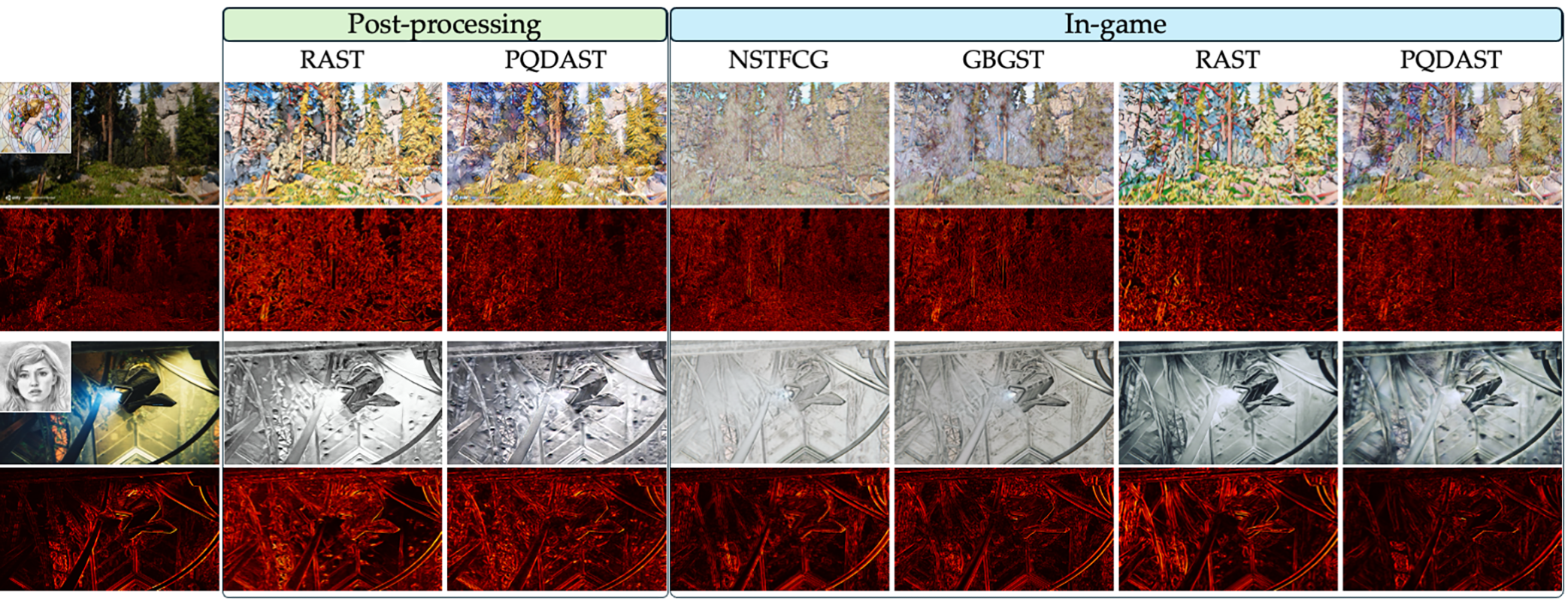}
   
     
  
    
   \caption{Results for additional game scenes/style images. The bottom rows provide the temporal error heatmap between the current and previous frame.} 
   \label{fig:results_comparison_2}
\end{figure*}

Qualitative results are shown in Figure~\ref{fig:qualitative}. The original input frame and style image are shown on the top left, whilst the bottom rows provide a temporal error heatmap showing the difference between the current and previous frame. It is important to note that specific representative frames and their corresponding heatmaps are provided in Figure~\ref{fig:qualitative} (and also in Figure~\ref{fig:results_comparison_2}). The error heatmaps would vary if other frames were chosen. A comprehensive quantitative evaluation for multiple frames and from different scenes is given in Section~\ref{sec:quantiative_results}. 
As shown in Figure~\ref{fig:qualitative}, \textit{CSBNet} and \textit{MCCNet} demonstrate good stylisation quality but with noticeable artefacts, mainly around the central object's edges. \textit{FVMST} does not adequately transfer the artistic style of the reference painting producing white blobs and inconsistent stylisations. \textit{AdaAttN} manages to retain important content information, however, the stylisation effect is not sufficiently achieved -- the stylised frame does not contain yellow colours that are eminent in the reference style. \textit{RAST} produces high-quality stylisations, justifying our selection for a `teacher' model to train our distillation framework. Nevertheless, the temporal error heatmap shows temporal incoherence. This is also noticeable for the other image and video approaches used as post-processing effects (\textit{AdaAttN}, \textit{CSBNet}, \textit{MCCNet}, \textit{FVMST}).
The in-game approaches \textit{NSTFCG} and \textit{GBGST} demonstrate improved temporal stability performance but sacrifice some stylisation quality. Our proposed system successfully compresses \textit{RAST}, maintaining a high degree of similarity to \textit{RAST} when used post-process while the corresponding temporal error heatmap is improved. When \textit{PQDAST} is used in-game, similar to \textit{NSTFCG} and \textit{GBGST}, the temporal error heatmap closely matches that of the input frame. Additionally, as our model distils knowledge from \textit{RAST}, the stylisation quality is considerably enhanced.

Additional qualitative comparisons are provided in Figure~\ref{fig:results_comparison_2}. The temporal error heatmap of \textit{PQDAST} (Post-processing) is similar to those of \textit{NSTFCG} and \textit{GBGST}, but the stylisation quality is noticeably better. \textit{PQDAST}'s heatmap more closely resembles the original frame's heatmap than \textit{RAST}'s temporal error heatmap does. Yet, there is a slight but noticeable difference in stylisations between our method and \textit{RAST}. This arises from the different types of data used for training. Our method is trained on synthetic frames, whereas \textit{RAST} was trained exclusively on real-world images from the MS COCO \cite{lin2014microsoft} dataset. This highlights the impact of dataset characteristics on the resulting stylistic outcomes, demonstrating the unique advantages and challenges presented by both synthetic and real-world datasets. In this work, we choose to utilise the conventional, widely used MPI Sintel \cite{Butler2012sintel} dataset to develop a system with broad generalisability across various games. This decision, instead of using the training set suggested in \cite{ioannou2024towards}, allows our model to adapt to arbitrary styles and different games. However, this approach comes at the cost of not being able to use G-buffer information. Thus we do not train our model solely using frames from the games that we test on, similarly to \cite{ioannou2024towards}.


\subsection{Quantitative Results}
\label{sec:quantiative_results}

\begin{table*}[htb]
    \caption{Quantitative results. Warping Error and LPIPS Error are both in the form $\times 10$. LPIPS measures perceptual similarity between original rendered frames and stylised frames. SIFID and ArtFID quantify the style performance. We provide results for our system, \textit{PQDAST}, injected in the pipeline and for the trained stylisation network applied as a post-process effect. We do the same for \textit{RAST}. The best results are indicated in \textbf{bold}, and the second best are \underline{underlined}.}
    \label{tab:results_temp_style}
    \begin{tabular}{l | c  c  c  c  c  c | c  c  c  c  c }
        \toprule
  
        & AdaAttN & CSBNet & MCCNet & FVMST & RAST  & PQDAST & NSTFCG & GBGST & RAST & PQDAST  \\
       
        \midrule        
        
        Warping Error~$\downarrow$ & 1.6477 & 1.7458 & 1.6519 & 1.8524 & 1.7119 & 1.6080 & 1.5798 & \underline{1.2984} & 1.4636 & \textbf{1.2695} \\
        LPIPS Error~$\downarrow$ & 0.3217 & 0.3908 & 0.3547 & 0.3215 & 0.5285 & 0.4730 & \underline{0.2930} & \textbf{0.2515} & 0.4131 & 0.3371 \\ 
        \midrule
        
        LPIPS~$\downarrow$ & \textbf{0.2692} & 0.3378 & 0.3468 & 0.3806 & \underline{0.3176} & 0.3294 & 0.3879 & 0.3494 & 0.3384 & 0.3327 \\
        SIFID~$\downarrow$ & 1.6115 & 2.2468 & \underline{1.5555} & 2.2529 & \textbf{1.2913} & 1.6185 & 1.8679 & 1.9401 & 3.4755 & 3.7163 \\
        ArtFID~$\downarrow$ & 49.4115 & 52.4232 & \underline{47.6695} & 53.8949 & \textbf{46.8992} & 51.3266 & 57.1858 & 54.1722 & 51.5226 & 52.8609 \\ 
        
          \midrule
        Processing Stage & \multicolumn{6}{c|}{Post-process} & \multicolumn{4}{c}{In-game} \\         
        
        \bottomrule
    \end{tabular}
\end{table*}

Evaluation in the field of style transfer remains an open problem. A range of computational metrics exist to quantify the performance of stylisation approaches, yet there is no standardised evaluation procedure, and the computational metrics utilised are reliable only to a certain degree \cite{ioannou2024evaluation}. Here, we show quantitative evaluation using a few metrics deemed to be the most relevant in the context of style transfer for computer games (Table~\ref{tab:results_temp_style}). For consistency with previous in-game stylisation methods \cite{ioannou2023games,ioannou2024towards}, we use the same test dataset of 2100 frames from 4 different game scenes, and the same 10 style images. Note that our trained model has not seen any frames resembling the test dataset during training. 

To gauge the effectiveness of our approach in producing temporally stable stylisations, we measure Warping Error using optic flow information. Similarly to \cite{ioannou2023games,ioannou2024towards}, we also calculate LPIPS Error \cite{zhang2018unreasonable}, as the average perceptual distances between consecutive frames. The results are gathered in Table~\ref{tab:results_temp_style}. Our method outperforms state-of-the-art approaches in Warping Error and performs competently in LPIPS Error. This shows that \textit{PQDAST} can generate artistically stylised results given any reference style image while sustaining temporal stability effectively.

In Section~\ref{sec:distillation}, we justify the use of \FLIP, which is utilised as an alternative to SSIM. As advised in \cite{nilsson2020understanding}, we avoid the use of SSIM for colour images. To measure how our approach performs in terms of perceptual and stylisation quality, we use LPIPS \cite{zhang2018unreasonable}, SIFID \cite{shaham2019singan}, and ArtFID \cite{wright2022artfid}. LPIPS gives a calculation of how well the perceptual information in the original frames is retained. SIFID is a measure of style fidelity, basically measuring FID for single images. ArtFID computes both the performance in capturing content information and reproducing the style image in a single metric. As depicted in Table~\ref{tab:results_temp_style}, while the performance of our proposed framework drops for the SIFID metric, our system performs competently in terms of retaining important content information (LPIPS), better than the previous in-game stylisation approaches (\textit{NSTFCG} \cite{ioannou2023games} and \textit{GBGST} \cite{ioannou2024towards}). Our method also outperforms the in-game methods in overall style transfer quality (ArtFID).

To measure the quality of the proposed knowledge distillation scheme, results are also included for \textit{RAST} -- we adapted \textit{RAST} and injected it into the game's pipeline in a similar way to \textit{PQDAST} to provide an additional comparison for our work. Our method trained on a synthetic video dataset with temporal considerations outperforms \textit{RAST} in terms of temporal consistency both when applying stylisations in the rendering pipeline and as a post-process effect. \textit{RAST} performs very well in perceptual similarity score (second best), but its performance drops when embedding it in the game. \textit{PQDAST}'s smaller size does not have a considerable impact when measuring LPIPS, and it outperforms \textit{RAST} when injected into the graphics pipeline. Similarly, while \textit{RAST} performs the best in SIFID and ArtFID metrics, our algorithm's effectiveness is competent, showing that compressing the transformer and the decoder does not result in substantial degradation of style transfer quality.

\begin{table}[htb]
\begin{center}
     \caption{Efficiency. Due to the complexity of the operations of the first four models in the table, they could not be exported to the appropriate format \cite{onnx_2019} for usage inside Unity game engine. Inference times (post-process) include inference through the \textit{VGG} network if necessary for extracting features used in stylisation. These are measured on a single Nvidia GeForce RTX 3090 GPU, with image resolution $1920 \times 1080$.}
    \label{tab:results_efficiency}
    \begin{tabular}{l c c c c}
        \toprule
        Method & No. Styles & Memory (MB) & Speed (ms) & fps \\
        \midrule
        AdaAttN & $\infty$ & 50.2 & 86.04 & - \\
        CSBNet & $\infty$ & 16.0 & 80.66 & - \\
        MCCNet & $\infty$ & 18.3 & 34.69 & -  \\
        FVMST & 120 & 18.0 & 17.39 & - \\ 
        NSTFCG & 1  & 3.03 & 50.21 & 10 \\
        GBGST & 1  & 4.19 & 54.01 & 10 \\ 
        RAST  & $\infty$  & 30.4 & 31.73 & 2 \\
        \textbf{PQDAST}  & $\infty$  & 15.7 & 26.06 & 5 \\
        \bottomrule      
    \end{tabular}
\end{center}
\end{table}


Table~\ref{tab:results_efficiency} provides efficiency analysis. A major advantage of \textit{PQDAST} compared to previous in-game stylisation methods is that it is capable of reproducing arbitrary styles. Although our method is faster when inference time is calculated outside Unity, it achieves approximately 5 fps in Unity. This can be justified by the number of inference runs required -- to compute a stylised image, 3 forward passes are needed: through the image encoder (VGG), through the transformer and through the decoder; \textit{NSTFCG} and \textit{GBGST} only require one forward pass. Our proposed framework, though, is significantly reduced in size compared to \textit{RAST} and is therefore faster.


\subsection{Ablation Study}

\subsubsection{Perceptual Quality-Guided Distillation Loss}

\begin{figure}[htb]
  \centering
   \includegraphics[width=\linewidth]{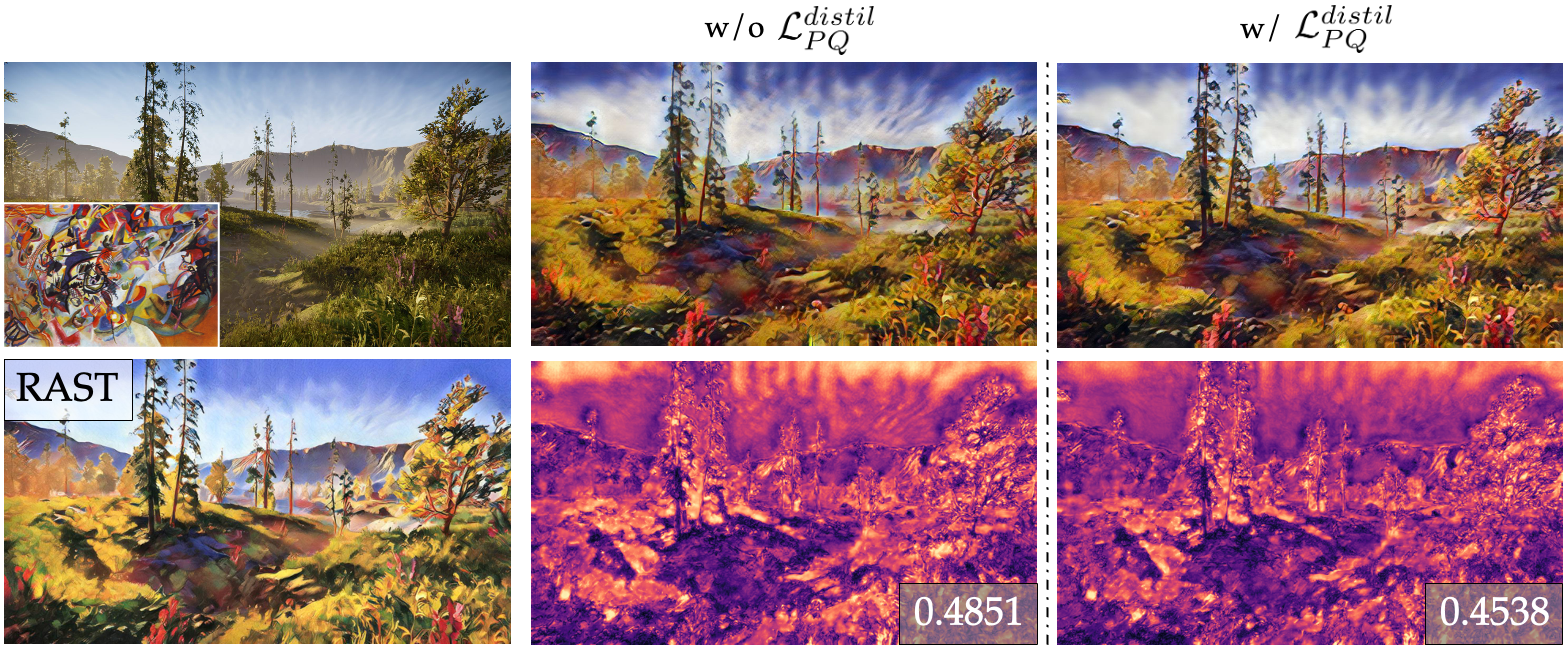}
   \caption{Ablation study on the effect of \FLIP for knowledge distillation. Using $\mathcal{L}_{PQ}^{distil}$ produces results that retain the detail of the content image (right) as in \textit{RAST}. Artefacts and inconsistencies are avoided (middle). The values from the \FLIP operation are shown at the bottom right of each difference image.}
   \label{fig:ablation_flip}
\end{figure}

Our proposed system synthesises results with comparable stylisation quality to \textit{RAST} (Figures~\ref{fig:results_ours_rast_2},~\ref{fig:qualitative}), justifying the effectiveness of using \FLIP in addition to matching the intermediate and output-level feature maps. To further gauge the effectiveness of \FLIP, we also train \textit{PQDAST} without $\mathcal{L}_{PQ}^{distil}$. Results are provided in Figure~\ref{fig:ablation_flip}. The difference between the generated result and \textit{RAST}'s generated result is also provided. Using \FLIP has a noticeable impact on the performance of our compressed model. Not only is the resulting image closer in similarity to \textit{RAST}, but it also avoids incongruities and uneven brushstrokes in parts of the image.

\subsubsection{Depth Loss}

\begin{figure}[htb]
  \centering
   \includegraphics[width=\linewidth]{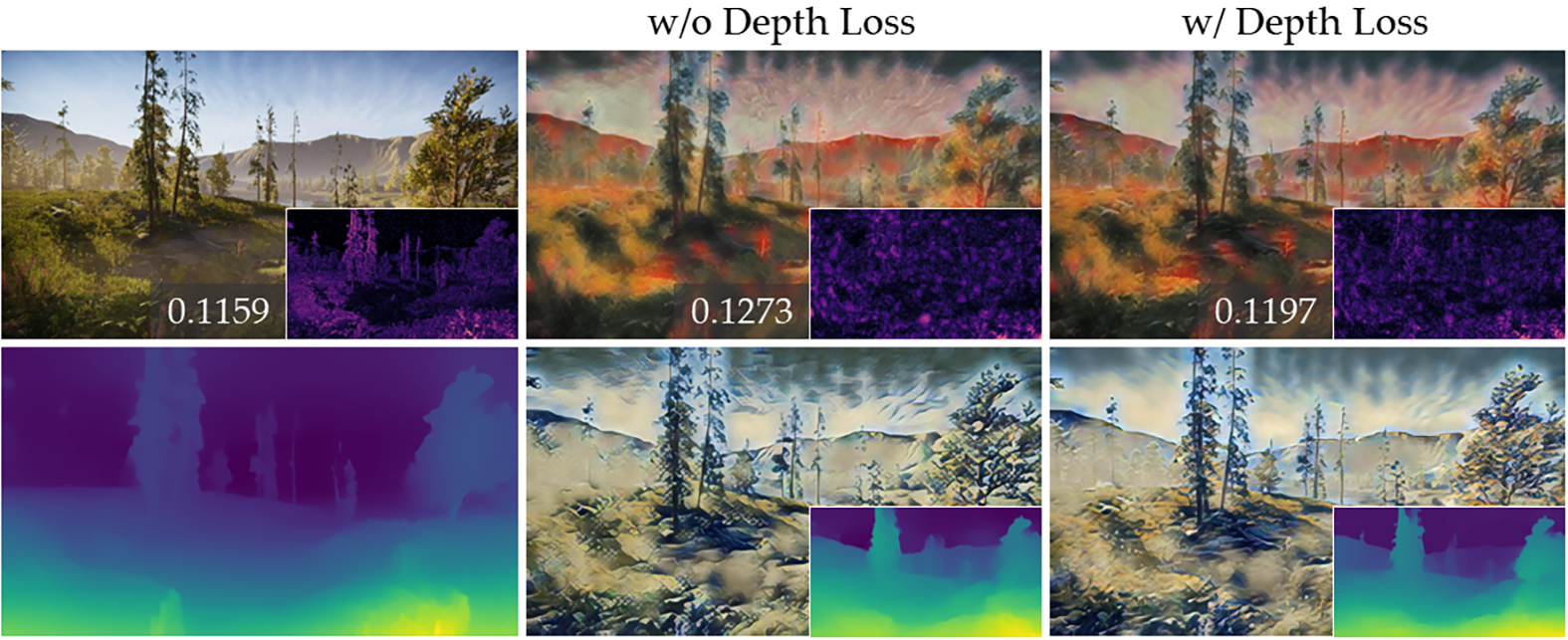}
   \caption{Ablation study on the effect of depth loss. The depth maps (bottom row) are generated using \textit{MiDaS} \cite{Ranftl2020}. The values from the \FLIP operation are shown on the left of each difference image in the top row.}
   \label{fig:ablation_depth}
\end{figure}

\begin{figure*}[htb]
  \centering
   \includegraphics[width=0.9\linewidth]{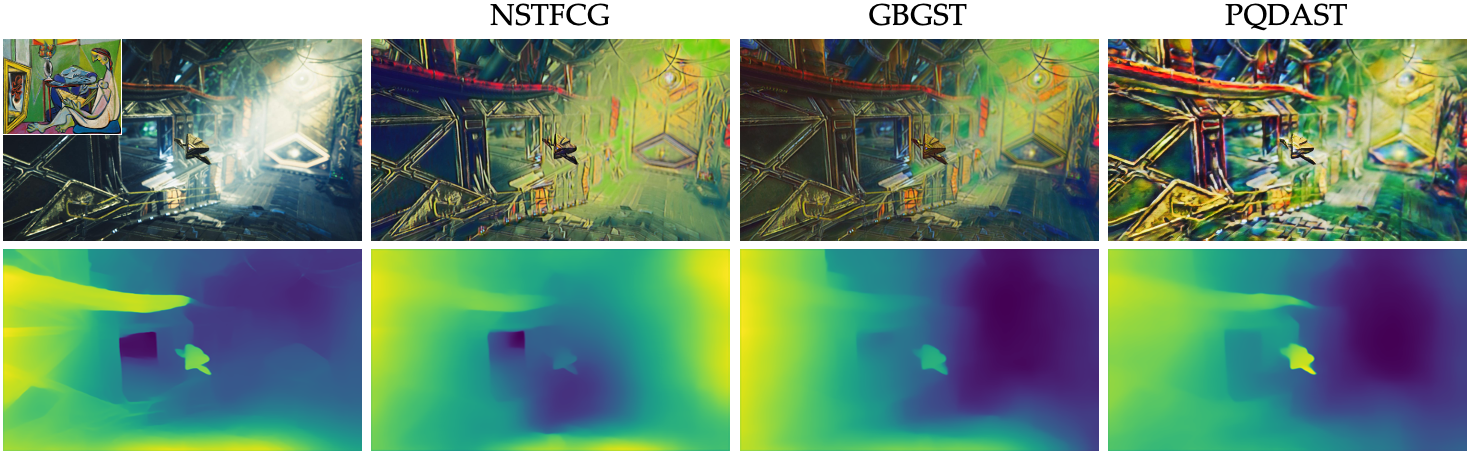}

    \hfill
    \centering
    
    \vspace{-0.25cm}
    \begin{tabular}{>{\centering\arraybackslash}m{3.95cm} >{\centering\arraybackslash} m{3.95cm}  >{\centering\arraybackslash} m{3.95cm}  }
    \toprule
    \textbf{NSTFCG} & \textbf{GBGST} & \textbf{PQDAST}  \\
    \midrule
      2092.80 / 14.9235 & 840.8185 / 18.8838 & 882.6262 / 18.6730   \\
            
      \bottomrule
    \end{tabular}

   \caption{Depth preservation performance comparison between our approach \textit{PQDAST (in-game)}, and the in-game methods \textit{NSTFCG} \cite{ioannou2023games}, \textit{GBGST} \cite{ioannou2024towards}. \textit{NSTFCG} and \textit{GBGST} use \textit{MiDaS} \cite{Ranftl2020} to define depth reconstruction loss. The depth maps in the bottom row are generated using \textit{MiDaS} for fairer comparisons. The table shows the error differences between the original frame's depth map and the depth map generated from the stylisation of each method. Mean square error (MSE) and peak-signal-to-noise ratio (PSNR) are provided.} 
   \label{fig:depth_comparison}
\end{figure*}

Employing a depth reconstruction loss has been established as a good practice for the style transfer task \cite{liu2017depth,ioannou2022depth,ioannou2023depth}. Compared to previous approaches, here, we have utilised an advanced depth prediction network that surpasses the performance of previously used methods. In Figure~\ref{fig:ablation_depth}, we show that incorporating the proposed depth loss has a noticeable impact not only in preserving depth and fine details (bottom row) but also in temporal consistency (top row -- temporal error maps are provided and the error computed is closer to that of the original frame). Our proposed framework, as injected in the game's pipeline, is also compared with other in-game approaches in Figure~\ref{fig:depth_comparison}. 
Although not trained using \textit{MiDaS}, our system's produced depth map is very similar to the original frame's depth map, demonstrating that \textit{PQDAST} preserves depth details and allows the main object in the centre of the frame to stand out. The calculated MSE and PSNR values illustrate that \textit{PQDAST} performs better than \textit{NSTFCG} and competently with \textit{GBGST} which also uses depth during inference. Employing an advanced depth prediction method, as discussed in \cite{ioannou2022depth}, results in better depth preservation.

\subsection{Limitations}

Despite our efforts to minimise temporal inconsistencies across sequential frames, preventing flickering and achieving temporal stability remains a challenge in the realm of games due to the complicated and unpredictable environments. Complex lighting and shadows often introduce flickering in the game scenes, even without post-processing effects taking place. Our approach aims to show that efficient artistic stylisation in games is possible without a large compromise in speed and memory. To further improve upon alleviating temporal or flickering issues, G-buffer information can be used at the inference stage, similarly to \cite{ioannou2024towards}. In this work we have not addressed that to avoid further inference delays.

As shown in Table~\ref{tab:results_efficiency}, although outperforming \textit{RAST}, the frame rate performance of our system drops to $\sim 5$ fps, whereas \textit{NSTFCG} and \textit{GBGST} achieve $10$ fps. However, these are capable of only one style per trained network. In the realm of computer games, speed remains an important issue for style transfer. Temporalisation schemes and manually scheduling the in-game network inference \cite{deliot_guinier_vanhoey_2020} could also help in improving speed. 

Another step towards better performance would be to compress the \textit{VGG} encoder used to generate encoded features. Notably, model compression can significantly impact the performance of artistically stylised games, offering advantages in both speed and GPU resource requirements. 
Unavoidably, speed is interconnected to the memory size of the model -- typically, a larger model would require more time to execute a forward pass. Additionally, although our stylisation model occupies a small amount of memory, it is important to note that memory constraints for games are significantly challenging. Modern games require more and more GPU memory, especially when targeting higher frame resolutions \cite{connatser2023gpu}. Our framework, the first to address arbitrary stylisation in games, avoids the dependence on multiple single-style-per-model NST models to reproduce multiple artistic styles, while our distillation algorithm promises a new way for compressing image generation models for use in games.


\section{Conclusion}
We have presented an arbitrary style transfer solution for computer games that makes use of a perceptual quality-guided knowledge distillation scheme inspired by image quality assessment of 3D renderings.
Our trained model, \textit{PQDAST}, is smaller and faster than the compared transformer-based arbitrary style transfer approaches, and it is integrated into the game's pipeline. Extensive qualitative and quantitative experiments have demonstrated that our system surpasses state-of-the-art methods in temporal coherence while achieving comparable perceptual and stylisation performance. Our work thus demonstrates an effective new way to perform knowledge distillation for image generation tasks, also showing that arbitrary style transfer for games can be achieved using a conventional GPU. Future work will focus on further improving the speed of the in-game stylisation models for real-time arbitrary style transfer in games.






\bibliographystyle{eg-alpha-doi} 
\bibliography{bib/references}       


\clearpage

\end{document}